\documentclass[10pt,twocolumn,letterpaper]{article}

\usepackage[pagenumbers]{cvpr} 

\usepackage{graphicx}
\usepackage{amsmath}
\usepackage{amssymb}
\usepackage{booktabs}
\usepackage{multirow}
\usepackage{multicol}
\usepackage[ruled,vlined]{algorithm2e}
\usepackage[T1]{fontenc}
\usepackage[misc]{ifsym}

\usepackage[pagebackref,breaklinks,colorlinks]{hyperref}

\usepackage[capitalize]{cleveref}
\crefname{section}{Sec.}{Secs.}
\Crefname{section}{Section}{Sections}
\Crefname{table}{Table}{Tables}
\crefname{table}{Tab.}{Tabs.}

\def\cvprPaperID{1887} 
\def\confName{CVPR}
\def\confYear{2022}

\begin{document}
\renewcommand{\thefootnote}{\fnsymbol{footnote}}
\title{Training BatchNorm Only in Neural Architecture Search and Beyond}

\author{Yichen Zhu$^{1}$\footnotemark[2], Jie Du$^{1,2}$\footnotemark[2] , Yuqin Zhu$^{1}$, Yi Wang$^{1}$, Zhicai Ou$^{1}$, Feifei Feng$^{1}$ and Jian Tang$^{1}$\footnotemark[4]\\
$^{1}$Midea Group, AI Innovation Center \\
$^{2}$Tongji University
}

\maketitle
\footnotetext[2]{Equal contribution.}
\footnotetext[4]{Corresponding author.}
\renewcommand{\thefootnote}{\arabic{footnote}}
\begin{abstract}

This work investigates the usage of batch normalization in neural architecture search (NAS). Specifically, Frankle \textit{et al.}~\cite{frankle2021training} find that training BatchNorm only can achieve non-trivial performance. Furthermore, Chen \textit{et al.}~\cite{chen2021bnnas} claim that training BatchNorm only can speed up the training of the one-shot NAS supernet over ten times. Critically, there is no effort to understand 1) why training BatchNorm only can find the perform-well architectures with the reduced supernet-training time, and 2) what is the difference between the train-BN-only supernet and the standard-train supernet. 

We begin by showing that the train-BN-only networks converge to the neural tangent kernel regime, obtain the same training dynamics as train all parameters theoretically. Our proof supports the claim to train BatchNorm only on supernet with less training time. Then, we empirically disclose that train-BN-only supernet provides an advantage on convolutions over other operators, cause \textbf{unfair competition} between architectures. This is due to only the convolution operator being attached with BatchNorm. Through experiments, we show that such unfairness makes the search algorithm prone to select models with convolutions. To solve this issue, we introduce fairness in the search space by placing a BatchNorm layer on every operator. However, we observe that the performance predictor in Chen \textit{et al.} is inapplicable on the new search space. To this end, we propose a novel composite performance indicator to evaluate networks from three perspectives: \textbf{expressivity}, \textbf{trainability}, and \textbf{uncertainty}, derived from the theoretical property of BatchNorm. We demonstrate the effectiveness of our approach on multiple NAS-benchmarks (NAS-Bench-101, NAS-Bench-201) and search spaces (DARTS search space and MobileNet search space).

\end{abstract}

\section{Introduction}
The big leap of convolutional neural networks (CNNs) starts from ResNet~\cite{he2016deep} and batch normalization~\cite{ioffe2015batchnorm}. It was initially considered as a technique to reduce internal covariance shift; however, recent works prove that it could smooth the loss landscape~\cite{santurkar2018bnsmooth},  stabilize the training  process~\cite{bjorck2018understandingbn, wu2021rethinkingbatch}, preserve rank stability under certain assumption~\cite{daneshmand2020batchrank}. Though some studies show that BatchNorm may be the cause of gradient explosion in CNNs~\cite{yang2018mean}, there are irreplaceable advantages of utilizing BatchNorm in CNNs.

Nevertheless, BatchNorm\footnote{For notation simplicity, we use both BatchNorm and BN interchangeably to represent the batch normalization layers.} is problematic in Neural Architecture Search (NAS) due to inaccurate batch statistics across sub-networks~\cite{elsken2021bag}, the small batch size on downstream tasks~\cite{chen2019detnas}. Existing works solve these issues by finetuning~\cite{guo2020single, chu2019fairnas}, fixing the learnable parameters, replacing with ghost batch normalization~\cite{bender2018oneshot, hoffer2017ghostbn}/group normalization~\cite{wu2018groupnorm, wang2020nasfcos}, or removing entirely from training~\cite{xu2019autofpn}. 
\\
\\
\noindent
\textbf{Train-BN-only\footnote{In our context, train BatchNorm only means to fix all parameters at their initialization states, and update the parameters of BatchNorm during training.} NAS.} Frankle \textit{et al.}~\cite{frankle2021training} was the first study to empirically show a CNN that is trained with BatchNorm only can achieve apparently non-random performance on multiple image classification datasets, including ImageNet. Later works explore the train-BN-only networks on feature orthogonality at deep layers~\cite{daneshmand2021batchortho} and influence on transfer learning~\cite{kanavati2021partial}. Recently, Chen \textit{et al.}~\cite{chen2021bnnas} proposed to speed up the supernet training over ten times by training the BatchNorm only and achieve comparable results on a modified MobileNet search space than state-of-the-art one-shot NAS algorithms. 
\\
\\
\noindent
This breakthrough shows the great potential of BatchNorm in NAS; meanwhile, it highlights two central questions: 1) Can we reduce the training time if only BatchNorm is trained? Intuitively, without updating the parameters in the networks other than BatchNorm requires less training time. However, unlike the conventional image classification task as in Frankle \textit{et al.}~\cite{frankle2021training}, it is crucial to preserve the rank consistency between various architectures in the supernet if the train-BN-only strategy can be successfully applied to NAS. Therefore, understanding the training dynamics~\cite{wilson1997training}, which reflects the optimization process of training neural networks and articulates models' generalization ability of train-BN-only networks, is essential. In this work, we demonstrate that, under certain conditions, the train-BN-only networks converge to the fixed neural tangent kernel regime~\cite{cntk, jacot2019freezeandchaos}. It reveals that the train-BN-only networks obtain the same training dynamic as networks train all parameters, which partially indicates BN-NAS's success in the NAS domain and supports the method to train BatchNorm only in NAS.

It comes up with the second question, 2) What is the difference between train-BN-only supernet and standard-train supernet? Typically, only the convolutions are attached with a BatchNorm. This is problematic since many predefined search spaces in NAS contain multiple operators other than convolutions (i.e., pooling, identity, etc.). These operators do not have a BatchNorm layer follow-up. In this paper, we disclose an unfair competition between architectures that are triggered by train-BN-only strategy. Because the BatchNorm has learnable parameters, an unfair advantage is established for the convolution operator (Detailed experiments are in \S\ref{sec:unfair}).



\begin{figure}[t]
     \centering
     \includegraphics[height=8.2cm, width=0.5\textwidth]{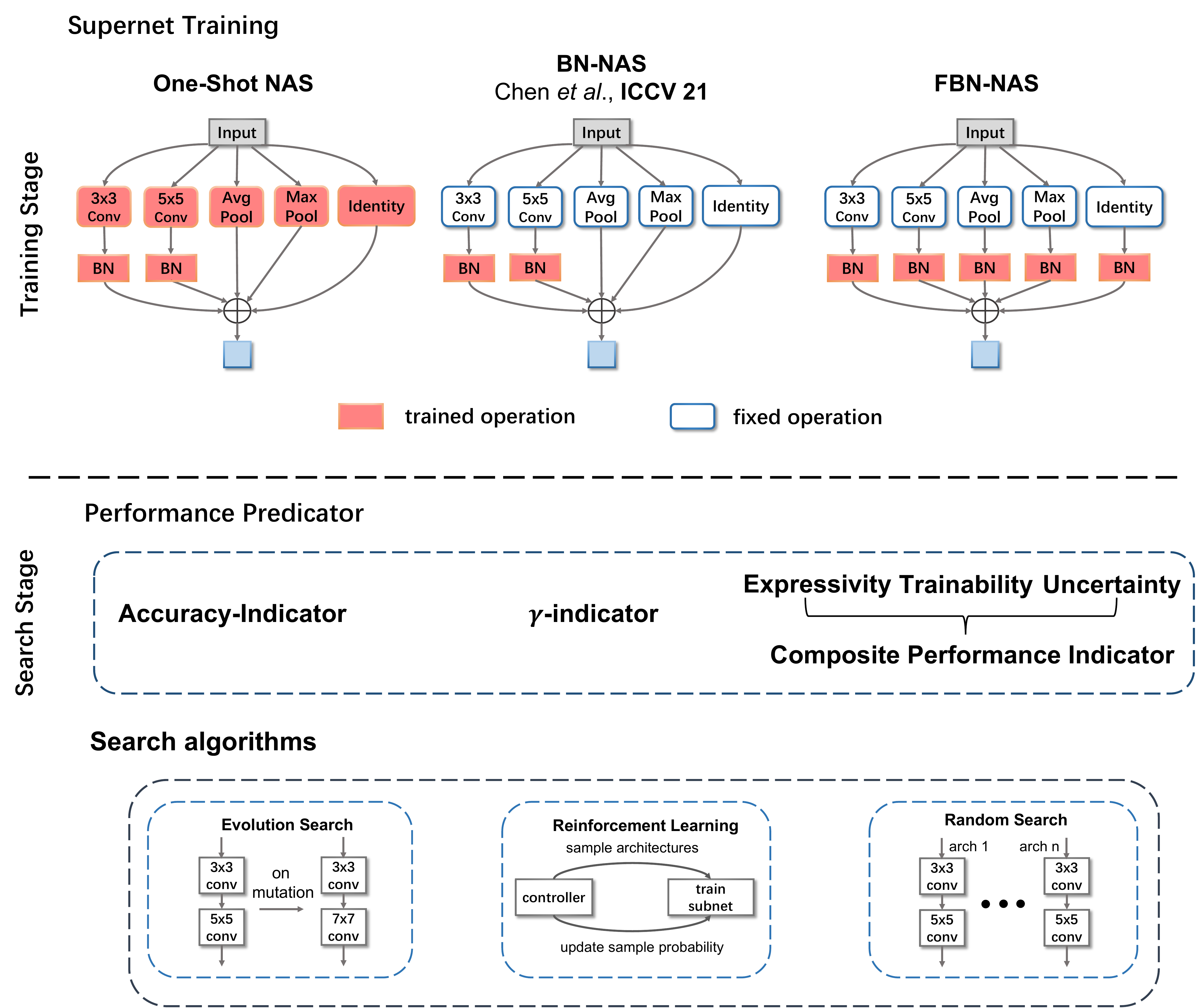}
     \caption{The comparison of FBN-NAS with BN-NAS and conventional one-shot NAS. Our approach use a new search space, and leverage theoretically inspired properties of BatchNorm as performance indicator. Best view in color.}
     \label{fig:overview}
\end{figure}

To remedy the unfairness, we place a BatchNorm layer after every operator, such that all architectures get the same chance to compete on the stage. Nevertheless, we empirically find that the previously effective performance predictor, \textit{gamma value}~\cite{chen2021bnnas} in BatchNorm, is no longer available after we bring BN fairness into the search space. Therefore, we need to design a new performance predictor that can efficiently and effectively evaluate the networks. As aforementioned, the BatchNorm brings many theoretically sound properties into the CNNs; thus, we leverage these properties to present a BatchNorm-based, theoretically-inspired composite performance predictor, which decomposes the generalization of neural networks to three perspectives: expressivity, trainability, and uncertainty. Notably, although the prior two properties have been discussed in training-free NAS, this is the first time in the literature to leverage BatchNorm to evaluate these two properties of networks in NAS. Moreover, to the best of our knowledge, the relationship between performance and model uncertainty has not been studied before. Our composite performance predictor shows a statistically significant correlation with model accuracy, verified on NAS-Bench-201. It helps our method find better and faster architecture in the search space than BN-NAS, while the training cost is further reduced. In Figure~\ref{fig:overview}, we demonstrate the overview of our approach and compare it to the previous one-shot NAS. We also show that our method enables faster supernet training and searches for more efficient and powerful models in various search spaces. 

In summary, our contributions are the following:
\begin{itemize}
    \item We theoretically prove that train-BN-only networks obtain the same training dynamic as training regular networks from the perspective of the neural tangent kernel. Our proof lay the foundation of applying the train BatchNorm only strategy to NAS.
    \item We empirically identify an unfair competition between architectures that is the consequence of the train-BN-only strategy. Therefore, we propose to insert BN after every operator to fix the issue. 
    \item Our work is the first  to leverage BatchNorm's theoretical property as the performance indicator to NAS. To the best of our knowledge, it is also the first time to link network performance with model uncertainty. 
\end{itemize}
\section{Related Works}
\label{sec:related_works}
\noindent
\textbf{Batch Normalization.}
Though many normalization methods~\cite{wu2018groupnorm, ba2016layernorm} have been proposed in the last few years, batch normalization~\cite{ioffe2015batchnorm} still dominates the CNNs. It was initially considered as a technique to reduce internal covariance shift~\cite{ioffe2015batchnorm}; however, recent works prove that it could smooth the loss landscape~\cite{santurkar2018bnsmooth}, stabilize the training process~\cite{bjorck2018understandingbn, wu2021rethinkingbatch}, and preserve rank stability under certain assumption~\cite{daneshmand2020batchrank}. BatchNorm can also be used to measure the model uncertainty~\cite{teye2018bayesian}. ~\cite{frankle2021training} have found out that training BatchNorm only while fixing the convolutional layers at the initial state can also achieve performance that is much better from random guess. 
\\
\\
\noindent
\textbf{Neural Architecture Search.} The over-parameterizated neural networks have demonstrate strong generalization ability on various tasks yet hard to deploy on embedding devices, thus, many techniques have been proposed to accelerate the model inferences, including pruning~\cite{han2015deep, frankle2018lottery}, knowledge distillation~\cite{gou2021knowledge, hinton2015distilling, zhu2021student}, quantization~\cite{polino2018model}. Neural architecture search (NAS), one of the most popular model compression techniques, has been extensively studied in the last few years. The initial techniques focused on reinforcement learning~\cite{zoph2018learning, pham2018enas} and evolutionary search~\cite{maziarz2018evolutionary, pham2018enas}, one-shot NAS algorithms~\cite{bender2018oneshot} and predictor-based NAS algorithms~\cite{shi2020bridging, wang2019alphax, white2021bananas}. However, these NAS methods are not practical due to their extremely long training and search times. Many researchers have redirected their interest in designing efficient and reliable NAS algorithms. Among them, weight-sharing is the most popular and promising technique. 
Given a pre-defined search space, the weight-sharing approach constructs a supernet that consists of all potential models. Conventional one-shot weight-sharing approaches~\cite{dong2019one, guo2020single, chu2019fairnas} first train a supernet and then apply the search strategy to find the optimal architectures. Another approach is differentiable weight-sharing~\cite{darts, cai2018proxylessnas, chu2020fairdarts}, which places a differentiable parameter for each node and updates alone with the architecture weights. The differentiable weight-sharing NAS is typically faster than one-shot NAS, yet hard to put hard constraints (i.e., FLOPs or latency) on the target model.

\section{Analysis}
\label{sec:whybnnas_biased}
\subsection{Train-BN-only Networks are Neural Tangent Kernel}
We first formally defined a fully-connected network with batch normalization layer. Consider a L-layers network with width $n_{l}$, for $l = 0, \cdots, L$. We denote the weight and bias for the $l$-th fully-connected layer by $\mathcal{W}^{l, fc}\in\mathcal{R}^{n_{l}\times n_{l-1}}$ and bias $b^{l, fc} \in \mathcal{R}^{n_{l}}$. The weight, bias, mean statistics for batch, variance statistics for batch for the $l$-th batch normalization layer attached to the $l$-th fully-connected layer is defined by $\mathcal{W}^{l, bn}\in\mathcal{R}^{n_{l}\times n_{l-1}}$, bias, mean and variance $b^{l, bn}, \mu^{l}, {\sigma^{2}}^{l} \in \mathcal{R}^{n_{l}}$. And a Lipschitz non-linearity function $\phi: \mathcal{R} \rightarrow \mathcal{R}$. For each input $x \in \mathcal{R}^{n_{0}}$, we denote the pre-activations by $h^{l}(x) \in \mathcal{R}^{n_{l}}$ and post-activations by $x^{l}(x) \in \mathcal{R}^{n_l}$. The feed-forward propagation for $l$-th layer in the network is
\begin{equation}
    x^{l}_{i} = \phi(\sum_{n_{l}}^{j=1} \mathcal{W}^{l, bn}_{ij}\frac{h_{i}^{l} - \mu^{l}_{i}}{\sqrt{{\sigma^{2}}^{l}_{i} + \epsilon}} + b^{l, bn}_{i})
\end{equation}
where
\begin{equation}
    h^{l}_{i} = \sum_{n_{l}}^{j=1} W^{l, fc}_{ij}x^{l-1}_{j} + b^{l, fc}_{i}
\end{equation}
The Neural Tangent Kernel (NTK) is proposed to characterize the behavior of fully-connected/convolutional infinite width neural networks whose layers have been trained by gradient descent. According to Jacot \textit{et al.}~\cite{jacot2018neuraltangentkernel}, the NTK of a fully-connected network with Gaussian initialization stays asymptotically constant during gradient descent training in the infinite-width limit, providing a guarantee for loss convergence. We find that the NTK of fixed fully-connected networks attached with trainable batch normalization have the same property in an asymptotic way.
\\
\\
\noindent
\textbf{Theorem 1.} \textit{Consider a FCN-BN of the equation (1) and (2) at Gaussian initialization $\mathcal{N}(0, 1)$, with a Lipschitz non-linearity $\sigma$, and the $\mathcal{W}^{l, fc}$ and $b^{l, fc}$ is fixed at initialization. In the limit as the layers width $n_{1}, \cdots, n_{L-1} \rightarrow \infty$, the NTK $\Theta^{L}_{0}(x, x')$, converges in probability to a deterministic limiting kernel:}
\begin{equation}
    \Theta^{L}_{0}(x,x') \rightarrow \Theta^{L}_{\infty}(x,x') \otimes \mathbf{I}_{n_{L} \times n_{L}}
\end{equation}
\textit{The scalar kernel $\Theta^{L}_{\infty}(x,x')$ is defined recursively by}
\begin{equation}
    \Theta^{l}_{\infty}(x,x') = \sigma^{2}_{w}\Sigma^{l}(x,x')\Theta^{l-1}_{\infty}(x,x') + \Sigma^{l}(x,x')
\end{equation}
\textit{where} 
\begin{equation}
    \Theta^{1}_{\infty}(x,x') = \Sigma^{l}(x,x') = \mathcal{E}_{f\sim\mathcal{N}(0,1)}[\phi(f(x))\phi(f(x'))]
\end{equation}
We refer readers to the Appendix for the derivation of this formula. Our theoretical result indicates train-BN-only networks (infinite-width networks of fixed, Gaussian initialization fully-connected layer with trainable Gaussian initialized BatchNorm layer) and standard trained networks (infinite-width networks of Gaussian initialization fully-connected layer) should have the same convergence rate during the gradient descent training. This means that two different training strategies have similar training dynamics in the NTK regime. Our analysis theoretically supports the practicability of the train-BN-only supernet. Additionally, since the training parameters in the train-BN-only setting is much less than standard-trained supernet, it is practicable to reduce the supernet training time. 

\subsection{Train-BN-only Supernet is Unfair} \label{sec:unfair}
\begin{figure}[t]
     \centering
     \centering
     \includegraphics[height=5.6cm,width=0.45\textwidth]{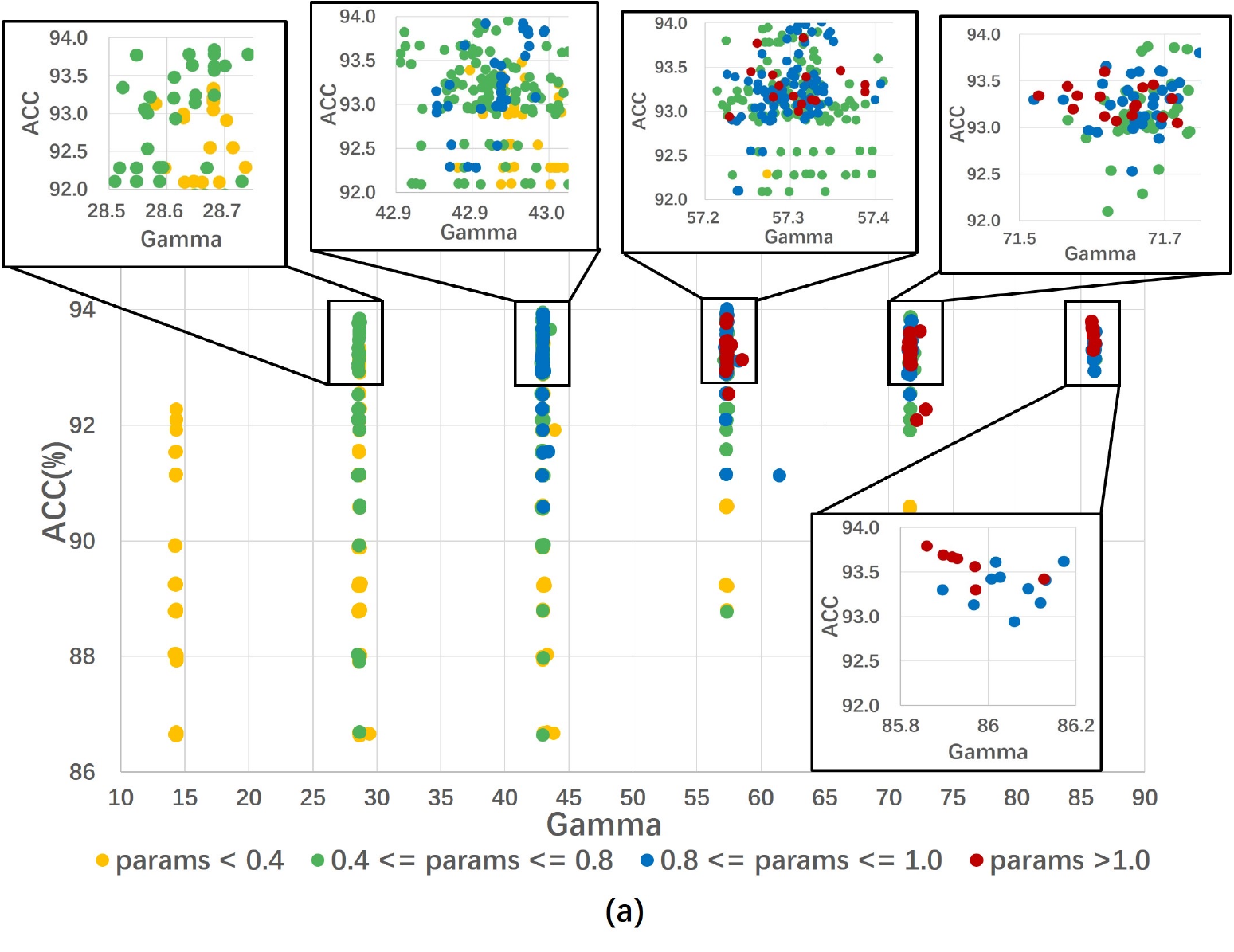}
     \caption{Analysis of BN-NAS and test accuracy on CIFAR10. The size of the models that are measured by number of parameters. Best view in colors. \textbf{Conclusion:} the BNNAS are biased toward large models.}
     \label{fig:biased_bnnas_study}
\end{figure}

In this section, we empirically study the behavior of train-BN-only supernet. The conventional one-shot NAS approach needs to train a supernet constructed by all operation candidates $OP$. A typical NAS search space has many types of operations, for instance, $OP(\cdot) \in \{3x3\_conv, avg\_pool, \cdots, identity\}$. Notably, all parameters in the operation candidates, including the BatchNorm, are updated during training. Then, it uses validation accuracy to rank the architectures. Recently, Chen \textit{et al.}~\cite{chen2021bnnas} propose to update the BatchNorm only and fix the parameters of $OP(\cdot)$. The search stage uses the average $\gamma$ value over all cells to indicate model rank. It is noteworthy that they limit the choice of $OP(\cdot) \in Conv(\cdot)$ such that the search space contains convolutional operators only.
\begin{table*}[t]
  \centering
  \renewcommand*\arraystretch{1.15}
  \begin{tabular}{l|c|c|c|c|c|c|c|c}
  \hline
  Method & 1x1 Conv & 3x3 Conv & Avg Pool & Identity & Zero & Avg Params & Avg FLOPs & Avg Acc \\
  \hline
  BN-NAS & 24 & 36 & 0 & 0 & 0 & 0.87 & 126 & 93.24 \\
  FBN-NAS & 16 & 26 & 4 & 11 & 3 & 0.69 & 103 & 93.28 \\
    \hline
  \end{tabular}
  \caption{We count the number of various operations for the Top-10 architecture with training stand alone architectures in NAS-Bench-201 on CIFAR-10. We present the result of BN-NAS and FBN-NAS.}
  \label{tbl:ops_number_counting}
\end{table*}
Figure~\ref{fig:biased_bnnas_study} shows the scatter plot of trained model test accuracy versus the value of gamma in BatchNorm. We randomly sample 1,000 models from NAS-Bench-201 and train BatchNorm only with each model by one epoch on CIFAR10. Each color represents a range of the number of parameters.

We can observe that there is indeed a positive correlation between test accuracy and gamma value. However, the architectures with a high gamma value (over 85) mostly have a more significant number of parameters (we show the corresponding plot measured by FLOPs in Appendix). Meanwhile, the test accuracy of these architectures is mostly comparable or even worse than many smaller architectures with low gamma values. In Table~\ref{tbl:ops_number_counting} we count the number of operations in the architectures with the top ten highest gamma values in the BN-NAS. We find out that the BN-NAS favors convolutional layers: None of the architecture with high gamma values have operations other than 1x1 and 3x3 convolutions. 

Our empirical observation unveils the fact that the train-BN-only supernet is biased towards convolution layers because \textit{the BatchNorm only attached to the convolutions}. In other words, since only the BatchNorm is trained, the capacity of networks is primarily determined by the learning process on the learnable parameters in BatchNorm. Even though we calculate the average value of gamma overall operation nodes, placing BatchNorm after convolutional layers only has a negative impact on searching good-performing architectures with small sizes. In conclusion, the BNNAS is biased, causing unfairness on the search in supernet.

\begin{figure}[t]
     \centering
     \begin{subfigure}[b]{0.5\textwidth}
         \centering
         \includegraphics[height=3.3cm,width=0.45\textwidth]{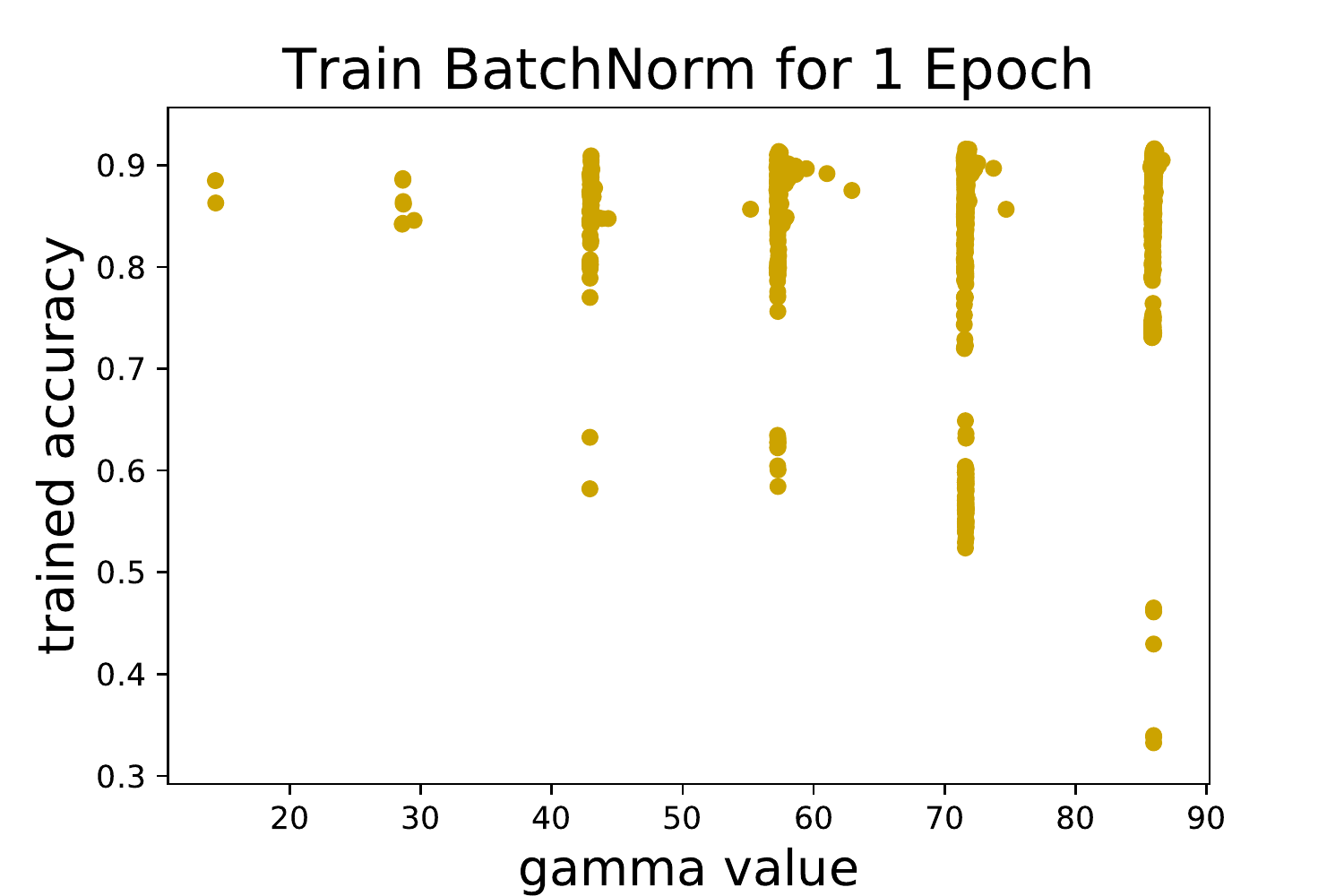}
         \centering
         \includegraphics[height=3.3cm,width=0.45\textwidth]{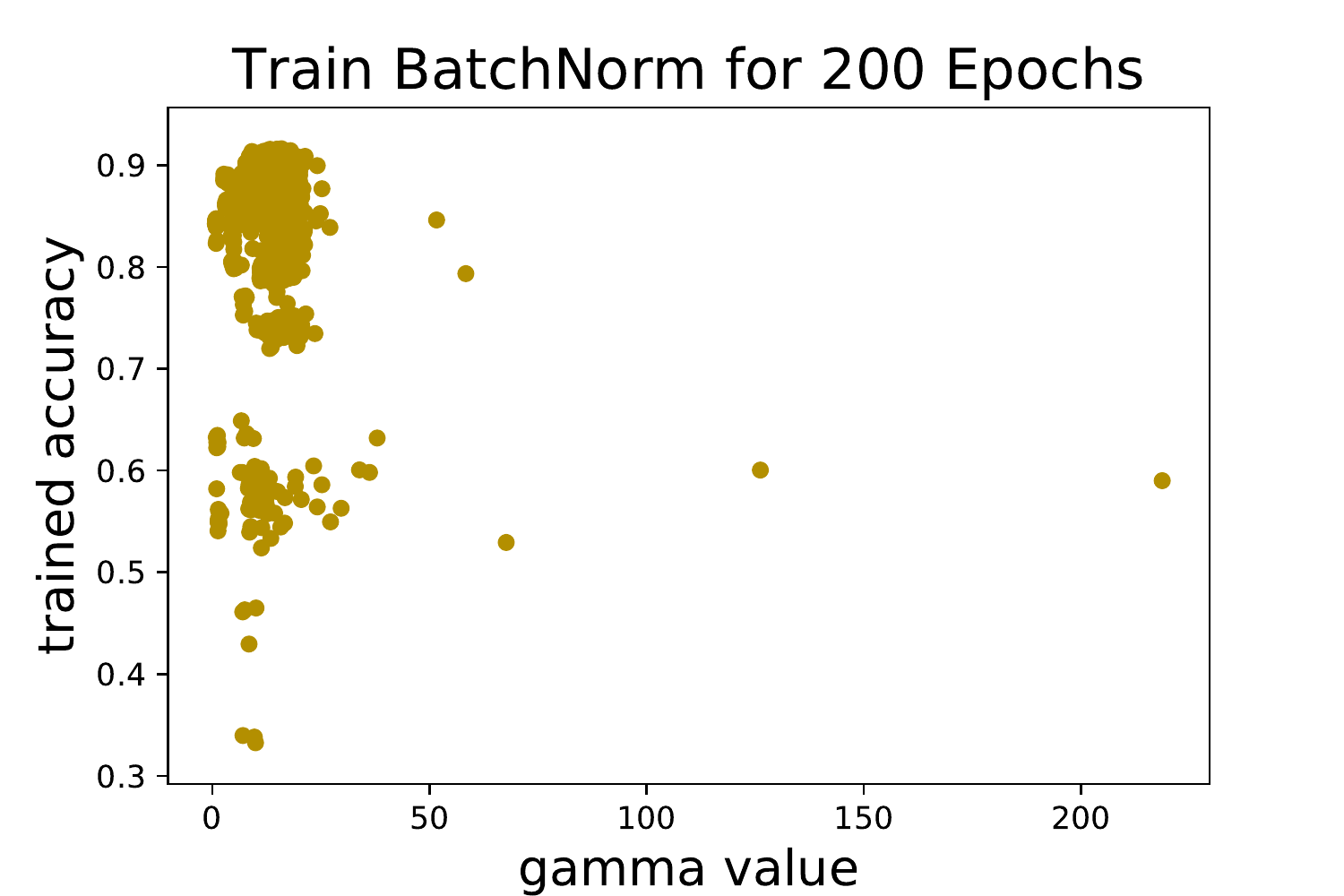}
     \end{subfigure}
     \caption{The gamma value of BatchNorm versus test accuracy train with 1 (left) and 200 (right) epoch. \textbf{Conclusion:} the $\gamma$-Indicator is \textit{ineffective} on the \textit{unbiased} BNNAS.}
     \label{fig:gamma_ind_notworking}
\end{figure}

\subsection{Fair-BN Supernet}
As aforementioned, fixing the unfairness issue in the search space for the Train-BN-Only supernet is necessary. Thus we propose the Fair BatchNorm supernet (FBN supernet). The method is simple: we attach a BatchNorm layer to every operation, including the pooling layer, the identity layer, etc. As such, a fair evaluation for each architecture can be done when the training is finished. This is extremely important for weight sharing NAS~\cite{chu2020darts-, liang2019darts+, chu2020fairdarts, chu2019fairnas}. The top part in Figure~\ref{fig:overview} gives a comparison of the BN-NAS and the unbiased FBN-NAS in the training stage. We also show the number of operations on the top ten architectures search by the unbiased FBN-NAS in Table~\ref{tbl:ops_number_counting}. Evidently, our approach can find networks consisting of diverse operations. While obtaining slightly better accuracy than the BNNAS, our approach consistently finds smaller architecture (measure by both FLOPs and number of parameters.).

\subsection{The Ineffectiveness of $\gamma$-indicator}
The most straightforward way to search architectures is to reuse the $\gamma$-indicator in BN-NAS~\cite{chen2021bnnas}. However, our empirical study shows that the correlation between test accuracy and gamma value no longer exists after the training becomes fair for various network architectures. The  Figure~\ref{fig:gamma_ind_notworking} shows the results of stand-alone training networks on CIFAR10 with training BN only for one epoch and two hundred epochs. There are 1,000 architectures randomly sampled from NAS-Bench-201. We can observe that when the networks are trained for only one epoch, the correlation between test accuracy and gamma value is completely random. The architectures with high gamma value can have an accuracy range from 30\% to 93\%. When we train the networks for 200 epochs, the situation becomes worse. In the right Figure~\ref{fig:gamma_ind_notworking}, the architecture that obtains the highest gamma value over 200 achieves around 60\% test accuracy. As we attach BatchNorm after pooling layers or identity layers, the gamma value of network architectures with those operations can also reach a very high value, which makes the $\gamma$-indicator no longer be an effective performance indicator for FBN supernet.

\section{The BatchNorm Based, Theoretically Inspired Composite Performance Indicator for NAS}
In the previous section, we have shown the convergence rate of train-BN-only supernet and the unfairness issue with the search process. We also validate that the $\gamma$-indicator fails on the Fair-BN supernet. In this section, we introduce a composite performance indicator (CPI) for the FBN supernet. Our goal is to design an indicator that can be 1) effectively predict the rank of the architectures \textit{at the early training stage of the unbiased BNNAS}, and 2) efficiently calculated during the search stage. We disentangle the model generalization into model expressivity, trainability, and uncertainty. More importantly, since FBN supernet only involves training on BatchNorm, we derive all three sub-performance indicators from theoretically inspired properties from BatchNorm.
\\
\\
\noindent
\textbf{Expressivity.} The ability of deep networks that can compactly express highly complex functions over input space is viewed as one of the most critical factors to their success. The expressivity of deep networks has been widely studied theoretically~\cite{exponentialexpressivity, raghu2017expressive} and has been used on NAS to score networks with random features~\cite{mellor2021trainfreenas, chen2021tenas}. However, both ~\cite{chen2021tenas} and ~\cite{mellor2021trainfreenas} count the number of linear regions of randomly initialized networks, limiting their search space on networks with ReLU activation~\cite{numberoflinearregions}. Whereas current SOTA backbones highly rely on non-linear activation other than ReLU, such as Swish~\cite{mobilenetv3, tan2019efficientnet}.

This paper discusses the expressivity of random features for various layer choices, such as convolutional layers, pooling layers, identity, etc., with trainable batch normalization. The batch normalization can be considered as a reparameterization trick on the parameters space. Learning to scale and shift the initialized parameters does not increase the expressivity of the random feature itself. Therefore, we refer to the validation accuracy as the expressivity of the deep networks. The original BNNAS fails to measure the expressivity of random features due to unfair supernet training. In our unbiased BNNAS, since the number of learnable BatchNorm parameters are the same for each operation, we can adopt expressivity as a performance indicator. 
\\
\\
\noindent
\textbf{Trainability.} Besides the expressivity, whether a deep network can achieve good performance is determined by how effectively the optimizer can optimize it. Recent work~\cite{santurkar2018bnsmooth} shows the effect of BatchNorm on the Lipschitzness of the loss, which plays a crucial role in optimization by controlling the amount by which the loss can change when taking a training step~\cite{boyd2004convex, nesterov2003convextintroductory}. Their empirical and theoretical results indicate that the BatchNorm smooth the loss landscape, and the re-parametrization in BatchNorm makes the gradient of the loss more Lipschitz. 

Specifically, let us denote the loss of a network with BatchNorm as $\mathcal{L_{BN}}$ and the loss of the same network without BatchNorm as $\mathcal{L_{noBN}}$. Given the activation $\hat{y}_{j}$, and gradient $\nabla_{\hat{y}_{j}}\hat{L}$, we define the Lipschitzness of the loss as $||\nabla_{y_{j}}\mathcal{L}||$, \cite{santurkar2018bnsmooth} prove that we can measure the effect of BatchNorm on the Lipschitz by
\begin{equation} \label{eq1}
\begin{split}
||\nabla_{y_{j}}\hat{\mathcal{L}}||^2 \leq \frac{\gamma^2}{\sigma^2_{j}} (||\nabla_{y_{j}}\mathcal{L}||^2 &- \frac{1}{m} \langle1, \nabla_{y_{j}}\mathcal{L}\rangle^2 \\&- \frac{1}{\sqrt{m}} \langle\nabla_{y_{j}}\mathcal{L}, \hat{y}_{j}\rangle^2)
\end{split}
\end{equation}
As a result, the scale term $\frac{\gamma}{\sigma}$ in the inequality can be used to measure the flatness of the loss landscape, which we define as the trainability score in our context.
\begin{table*}[h]
  \centering
  \renewcommand*\arraystretch{1.15}
  \begin{tabular}{lccccccccc}
    \hline
    \multirow{2}{*}{Method}     &  Search  & \multicolumn{2}{c}{CIFAR-10}  & &\multicolumn{2}{c}{CIFAR-100} & &  \multicolumn{2}{c}{ImageNet-16-120}  \\
    \cline{3-4} \cline{6-7} \cline{9-10}
    & (s)& validation & test &  & validation & test & & validation & test \\
    \hline
    \multicolumn{10}{c}{\textbf{Search Space: NAS-Bench-201}} \\
    \hline
    \multicolumn{10}{c}{\textbf{Non-Weight Sharing}} \\
    REA~\cite{real2019regularizednas} & 12000 & 91.19$\pm$0.31 & 93.92$\pm$0.30 & & 71.81$\pm$1.12 & 71.84$\pm$0.99 & & 45.15$\pm$0.89 & 45.54$\pm$1.03\\
    RS~\cite{li2020randomsearch} & 12000 & 90.93$\pm$0.36 & 93.70$\pm$0.36 & & 70.93$\pm$1.09 & 71.04$\pm$1.07 & & 44.45$\pm$1.10 & 44.57$\pm$1.25\\
    REINFORCE~\cite{williams1992simple} & 12000 & 91.09$\pm$0.37 & 93.85$\pm$0.37 & & 71.61$\pm$1.12 & 71.71$\pm$1.09 & & 45.05$\pm$1.02 & 45.25$\pm$1.18\\
    BOHB~\cite{falkner2018bohb} & 12000 & 90.82$\pm$0.53 & 93.61$\pm$0.52 & & 70.74$\pm$1.29 & 70.85$\pm$1.28 & & 44.26$\pm$1.36 & 44.42$\pm$1.49\\
    \hline
    \multicolumn{10}{c}{\textbf{Weight Sharing}} \\
    WSRS~\cite{guo2020single} & 7587 & 84.16$\pm$1.69 & 87.66$\pm$1.69 & & 59.00$\pm$4.60 & 58.33$\pm$4.34 & & 31.56$\pm$3.28 & 31.14$\pm$3.88\\
    DARTS~\cite{darts}  & 29902 & 39.77$\pm$0.00 & 54.30$\pm$0.00 & & 15.03$\pm$0.00 & 15.61$\pm$0.00 & & 16.43$\pm$0.00 & 16.32$\pm$0.00\\
    GDAS~\cite{you2020greedynas} & 28926 & 90.00$\pm$0.21 & 93.51$\pm$0.13 & & 71.14$\pm$0.27 & 70.61$\pm$0.26 &  &41.70$\pm$1.26 & 41.84$\pm$0.90\\
    FairNAS~\cite{chu2019fairnas} & 9845 & 90.07$\pm$0.57 & 93.23$\pm$0.18 & & 70.94$\pm$0.94 & 71.00$\pm$1.46 & & 41.90$\pm$1.00 & 42.19$\pm$0.31 \\
    SETN~\cite{dong2019one} & 31010 & 82.25$\pm$5.17 & 86.19$\pm$4.63 & & 56.89$\pm$7.59 & 56.87$\pm$7.77 & & 32.54$\pm$3.63 & 31.90$\pm$4.07 \\
    ENAS~\cite{pham2018enas} & 13315 & 39.77$\pm$0.00. & 54.30$\pm$0.00 & & 15.03$\pm$0.00 & 15.61$\pm$0.00 & & 16.43$\pm$0.00 & 16.32$\pm$0.00 \\
    \hline
    \multicolumn{10}{c}{\textbf{Training-Free}} \\
    NASWOT~\cite{mellor2021trainfreenas} & 306.19 & 89.69$\pm$0.73 & 92.96$\pm$0.81 & & 69.86$\pm$1.21 & 69.98$\pm$1.22 & & 43.95$\pm$2.05 & 44.44$\pm$2.10\\
    TENAS~\cite{chen2021tenas} & 1558  &89.92$\pm$0.43 & 93.35$\pm$0.25 & & 69.25$\pm$0.71   &69.59$\pm$0.67 & & 43.55$\pm$2.54 & 44.06$\pm$2.19\\
    \hline
    \multicolumn{10}{c}{\textbf{Training-BN}} \\
    BN-NAS~\cite{chen2021bnnas}  & 967  & 89.72$\pm$0.42 & 92.62$\pm$0.31 & & 68.46$\pm$1.04 & 68.33$\pm$0.89 & & 38.97$\pm$1.02 & 39.22$\pm$0.80 \\
    \textbf{FBN-NAS} & \textbf{686} & \textbf{90.32$\pm$0.24} & \textbf{93.79$\pm$0.18} &  & \textbf{70.87$\pm$0.53} & \textbf{70.91$\pm$0.89} & & 
    \textbf{44.17$\pm$0.14} & \textbf{44.35$\pm$1.18} \\
    \hline
    Optimal & - & 91.61 & 94.37 & & 74.49 & 73.51 & & 46.77 & 47.31 \\
    \hline
  \end{tabular}
  \caption{The mean$\pm$std. accuracies on NAS-Bench-201. Baselines are run over 500 times for most cases and 3 for weight-sharing methods and average accuracy is reported.}
  \label{tbl:nas201_exp}
\end{table*}
\\
\\
\noindent
\textbf{Uncertainty.} Uncertainty is a measurement of knowing what the model does not know. Deep neural networks are known to be overconfident in their predictions. Instead of point estimation, the model can offer confidence bounds for each decision from the probabilistic view to avoid over-confidence. This is useful to solve core issues of deep networks such as poor calibration and data inefficiency~\cite{mcdropout, kendall2017uncertainties}. There are rarely studies that discuss the relationship between uncertainty and model generalization. In this work, we obtain practical uncertainty for each network architecture via trainable batch normalization~\cite{mcbn}. We discuss the correlation of uncertainty and performance, and we use it as one of the performance indicators to find plausible architectures. 

Specifically, given a network architecture $X$, we are interested in approximating inference in Bayesian modeling. A common approach is to learn a parameterized approximating distribution $q_{\theta}(\omega)$ that minimizes $KL(q_{\theta}(\omega)||p(\omega|D))$, where $D$ is the training set, $\omega$ is the model parameters, $p(\omega|D)$ is the probabilistic model, the Kullback-Leibler divergence is of the true posterior with respect to its approximation. 

For models that are built upon batch normalization, we are able to obtain approximate posterior by modeling $\mu_{B}$ and $\sigma_{B}$ over a mini-batch $B$ as stochastic variables. Following ~\cite{mcbn}, we use approximate posterior to express an approximate predictive distribution $p(y|x, D) = \int f_{\omega}(x, y)q_{\theta}(\omega)d\omega$, and we can obtain the covariance of the predictive distribution empirically by 
\begin{equation} \label{eq1}
\begin{split}
Cov_{p*}[y] &\approx \tau^{-1}I + \frac{1}{\tau}\sum^{T}_{i=1}f_{(\hat{\mu_{i}}, \hat{\sigma_{i}})}(x)^{T}f_{(\hat{\mu_{i}},  \hat{\sigma_{i}})}(x) \\
 &- E_{p*}[y]^{T}E_{p*}[y]
\end{split}
\end{equation}
The calculation of variance is generic in practice. We simply record the value of variance for each iteration during evaluation and then average the total variance to obtain the final uncertainty score.

\noindent
\subsection{Composite Performance Indicator}
In the previous section, we present three theoretically inspired performance indicator to evaluate the performance of networks in FBN supernet. The key to build an effective NAS algorithm is how to combine these indicators. Intuitively, the numerical value for each indicator varies, thus it is impossible to simply adding them together. The naive approach would be to normalize each indicator's value, then add them into a single score. However, such a way requires us to know the estimated range of the value prior to we perform the search, which is impossible for most cases. Therefore, we rank the architecture according to the value for each indicator instead. How to combine the rank results on different indicators can be treated as an ensemble learning problem. In practice, we give equivalent importance to each indicator and use the average rank across three indicators as the final rank for simplicity. We discuss the correlation between accuracy and the CPI in Section~\ref{sec:cpi_perform}.
\begin{table}[t]
  \centering
  \renewcommand*\arraystretch{1.15}
  \begin{tabular}{lcc}
    \hline
    Method & Accuracy & Search Cost \\
    \hline
    Random~\cite{li2020randomsearch} &  90.38$\pm$5.51 & N/A \\
    REA~\cite{rea} &  93.87$\pm$0.22 & 12000 \\
    AREA~\cite{mellor2021trainfreenas} &  93.91$\pm$0.29 & 12000 \\
    RLNAS~\cite{randomlabelnas} & 93.78$\pm$0.14 & 10372\\
    \hline
    BN-NAS~\cite{chen2021bnnas}    & 91.04$\pm$0.32 & 979 \\
    \textbf{FBN-NAS}      & \textbf{94.16$\pm$0.10} & \textbf{701}\\
    \hline
  \end{tabular}
  \caption{The mean$\pm$std. accuracy with search cost on NAS-Bench-101.}
  \label{tbl:nas101_results}
\end{table}
\section{Experiments}
In this section, we introduce the experiments on NAS-Bench-101, NAS-Bench-201, DARTS search space, and MobileNet search space. All search cost are evaluated on Nvidia GTX 1080-Ti. 
\subsection{NAS-Benchmark Experiments}
\noindent
\textbf{NAS-Bench-101.}~\cite{nas101} is the first reproductive benchmark for NAS, and we compare our FBN-NAS with random search~\cite{nas101}, REA~\cite{rea}, AREA~\cite{mellor2021trainfreenas}, and RLNAS~\cite{randomlabelnas}. As shown in Table~\ref{tbl:nas101_results}, our FBN-NAS outperforms the BN-NAS counterpart by a large margin, on an average accuracy of 94.16\% versus 91.04\%. Our approach also achieves slightly higher accuracy than other methods with much less search cost, for instance, 6.8\% of RLNAS and 5.8\% of non-weight sharing REA.
\\
\\
\noindent
\textbf{NAS-Bench-201.} The NAS-Bench-201 is proposed by ~\cite{nas201}, which benchmarks a number of NAS algorithms. It contains 15,625 candidate architectures and provides the accuracy for each architecture. We compare our proposed methods with non-weight sharing NAS (Random Search, REA~\cite{real2019regularizednas}, REINFORCE~\cite{williams1992simple}, BOHB~\cite{falkner2018bohb}), weight-sharing NAS (Random Search~\cite{li2020randomsearch}, DARTS~\cite{darts}, GDAS~\cite{you2020greedynas}, SETN~\cite{dong2019one}, ENAS~\cite{pham2018enas}, FairNAS~\cite{chu2019fairnas}, BN-NAS~\cite{chen2021bnnas}), and training-free NAS (NASWOT~\cite{mellor2021trainfreenas} and TENAS~\cite{chen2021tenas}). We report the experimental results of FBN-NAS in Table~\ref{tbl:nas201_exp}.  Our results indicate the FBN-NAS is extremely fast: over 20 to 30 times faster than the weight-sharing approach (FairNAS and ENAS), while we obtain comparable or slightly better performance than these methods. Compared to the BN-NAS, our approach further save 70\% of search time and achieve better performance on all three datasets. Though our method does not designed for training-free approach, we compare our method with two state-of-the-art training-free NAS. Due to the high query time~\cite{white2021powerful}, the TENAS spend longer search time than FBN-NAS, while the performance is lower than ours. The NASWOT is about 2 times faster in terms of search cost, but our results are much better than NASWOT.


\begin{table}[t]
  \centering
  \renewcommand*\arraystretch{1.15}
  \begin{tabular}{l|c|c|c}
    \hline
    \multirow{2}{*}{Method}     &  Acc  & FLOPs& Search Cost\\
       & (\%)  &  (M)   & (GPU days)  \\
    \hline
    \multicolumn{4}{c}{\textbf{Search Space: MobileNet}} \\
    \hline
    ProxylessNAS~\cite{cai2018proxylessnas} & 75.1   & 465 & 8.3  \\
    FBNet~\cite{wu2019fbnet}  &  74.9   & 375 & 9  \\
    AngleNAS~\cite{hu2020anglenas} & 75.97  & 472  & 10 \\
    FairNAS~\cite{chu2019fairnas}  & 74.07    & 325 & 16  \\
    SPOS~\cite{guo2020single}   &  75.73  & 470  &  12  \\
    \hline
    BNNAS~\cite{chen2021bnnas} (FairNAS)  &  74.12  & 326 & 0.8  \\
    \textbf{FBN-NAS (FairNAS)} & \textbf{74.28}   & \textbf{325} & \textbf{0.5} \\
    \hline    
    BNNAS~\cite{chen2021bnnas} (SPOS)  &  75.67   & 470 & 1.2  \\
    \textbf{FBN-NAS (SPOS)} &  \textbf{75.81}  & \textbf{465}  & \textbf{0.8} \\
    \hline
    BNNAS~\cite{chen2021bnnas} (SPOS) + SE &  76.78  & 473  & 0.8 \\
    \textbf{FBN-NAS (SPOS) + SE} &  \textbf{76.83}  & \textbf{327}  & \textbf{0.5} \\
    \hline
    \multicolumn{4}{c}{\textbf{Search Space: DARTS}} \\
    \hline
    DARTS~\cite{darts} & 73.3 & 574& 4\\
    GDAS~\cite{you2020greedynas} & 74.4 & 590 & 0.3\\
    PC-DARTS~\cite{xu2019pcdarts}  & 75.8   &  597 & 3.8  \\
    Fair-DARTS~\cite{chu2020fairdarts} & 75.1 & 541 & 0.4 \\
    RLNAS~\cite{randomlabelnas}  &  75.9   & 597  &  8   \\
    \hline
    BNNAS~\cite{chen2021bnnas} (SPOS)  & 74.7 & 598 & 0.5\\
    \textbf{FBN-NAS (SPOS)} & \textbf{75.7}  &  \textbf{534}  & \textbf{0.33} \\
    \hline
  \end{tabular}
  \caption{The Top-1 accuracy on ImageNet with MobileNet search space and DARTS search space.}
  \label{tbl:imagenet_results}
\end{table}
\subsection{ImageNet Experiments}
\noindent
\textbf{MobileNet search space.} MobileNet-like Search Space consists of 21 to-be-searched layers, each layer leverages the MobileNetV2 inverted bottleneck~\cite{sandler2018mobilenetv2} and optional with SE module~\cite{hu2018senet}. At each layer, the search method can choose operation between convolutional layers with kernel size \{3, 5, 7\}, expansion ratio \{3, 6\} and identity layers. Note that the original BN-NAS does not have the identity layer option. We compare with several weight-sharing NAS that use the same search space setting~\cite{guo2020single, chu2019fairnas, cai2018proxylessnas, chen2021bnnas, hu2020anglenas}. We developed our method based on SPOS~\cite{guo2020single} and FairNAS~\cite{chu2019fairnas}, following the BN-NAS~\cite{chen2021bnnas} for fair comparison.As shown in Table~\ref{tbl:imagenet_results}, with the same search space, our searched results achieve 74.28\% with 325 MFLOPs and 75.81\% with 465 MFLOPs, slightly outperform SPOS and FairNAS counterpart, whereas the search time is only 24 times and 20 times less the of the original method, respectively. Furthermore, there is a clear advantage of FBN-NAS over the BN-NAS, which outperforms the BN-NAS by 0.16\% and 0.14\% on accuracy with 1.6 times and 1.5 times less search cost. It indicates that even on the search space that is mainly composed of convolutions, our method is still superior to the BN-NAS. By searching the SE module, we obtain architecture with 1.05\% higher accuracy than the BN-NAS with only 69\% of the FLOPs. Our unbiased training strategy and performance indicator help us find small architectures with plausible accuracy.
\\
\\
\noindent
\textbf{DARTS search space.} We compare our algorithms with various NAS algorithms~\cite{chu2019fairnas, you2020greedynas, darts, randomlabelnas, chu2020fairdarts} that performed on the same DARTS search space. The bottom of Table~\ref{tbl:imagenet_results} provides the experimental results on ImageNet. We can observe that the BN-NAS does not achieve satisfactory results, as the test accuracy is relatively lower than state-of-the-art methods, yet its FLOPs are similar or higher. On the other hand, the FBN-NAS obtain 1.0\% higher accuracy and 64 fewer FLOPs than the BN-NAS. It demonstrates that our approach is robust on search space with various operations. This further backs up our initial assumption that the biased BN-NAS hurts search performance.
\begin{figure}[t]
     \centering
     \centering
     \includegraphics[width=0.5\textwidth]{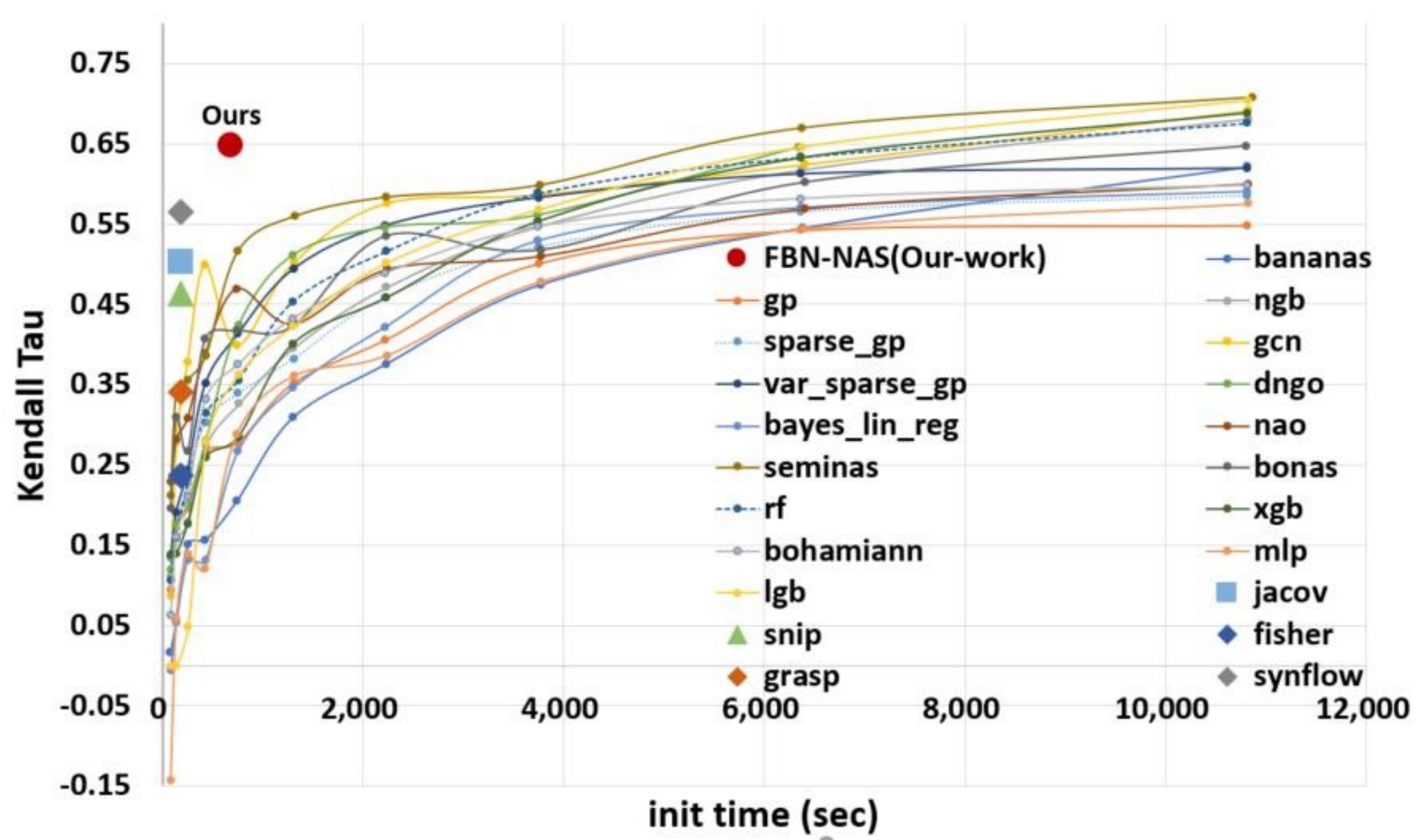}
     \caption{The correlation analysis of performance indicator on NAS-Bench-201 CIFAR100.}
     \label{fig:init_time_kendalltau}
\end{figure}

\subsection{Ablation Study}
\noindent
\textbf{Correlation analysis for composite performance indicator.} \label{sec:cpi_perform} One straightforward question for our approach is how strong is the correlation between the performance of networks with our composite performance indicator. Thus, we conduct a correlation analysis on NAS-Bench-201~\cite{nas201} CIFAR100. We follow White \textit{et al.} ~\cite{white2021powerful}, a through study on performance indicator in NAS. We compare our approach with a number of performance predictors, including training-free NAS (Jacobian Covariance in NASWOT~\cite{mellor2021trainfreenas}, Fisher, SNIP, Grasp, SynFlow in ZeroCostProxy~\cite{zerocost}) and many state-of-the-art methods~\cite{white2021bananas, rasmussen2003gaussian, siems2020bench, wen2020neural, bauer2016understanding, snoek2015scalable, titsias2009variational, luo2018neural, shi2020bridging, luo2020semi, siems2020bench, bishop2006pattern, springenberg2016bayesian,luo2020accuracy}.


As illustrated in Figure~\ref{fig:init_time_kendalltau}, our approach obtained the highest Kendall tau correlation under the same initialization time. Compared to the training-free NAS, our approach achieve a higher correlation with a short initial time overhead. In terms of initialization time, our method is highly efficient compared to the other performance predictor. Notably, the NAO~\cite{luo2018neural} needs to around eight more times for initialization in order to achieve a similar Kendall tau correlation. Overall, our composite performance indicator shows to be effective in ranking architectures.
\\
\\
\noindent
\textbf{Trainability-indicator versus $\gamma$-indicator.} Notice that even though the only difference between our trainability-indicator and $\gamma$-indicator is the term $\frac{1}{\sigma}$, the outcome is distinct. In Figure~\ref{fig:mobilenet_supernet_gamma}, we give a comparison between trainability-indicator and $\gamma$-indicator on a Mobilenet supernet, that is trained with unbiased FBN-NAS by 5 epochs on CIFAR-10. As expected, the value of $\gamma$-indicator is clustered together, which behaves similar to what we observe in previous study on training stand-alone networks. On the other hand, the trainability score scatter and shows a Kendall tau correlation of 0.646 with the test accuracy.
\begin{figure}[t]
     \centering
     \begin{subfigure}[b]{0.52\textwidth}
         \centering
         \includegraphics[height=3.5cm,width=0.49\textwidth]{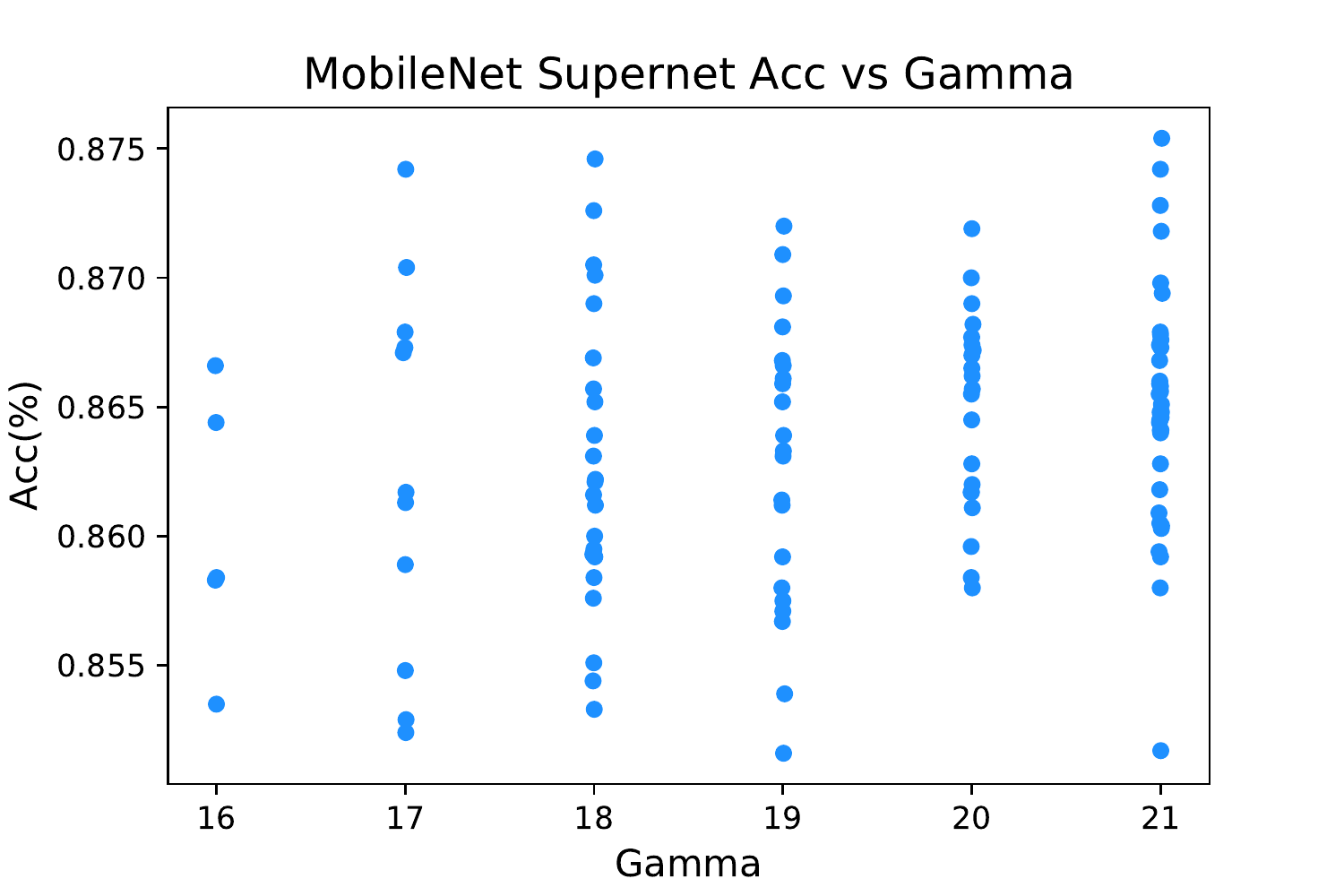}
         \centering
         \includegraphics[height=3.5cm, width=0.49\textwidth]{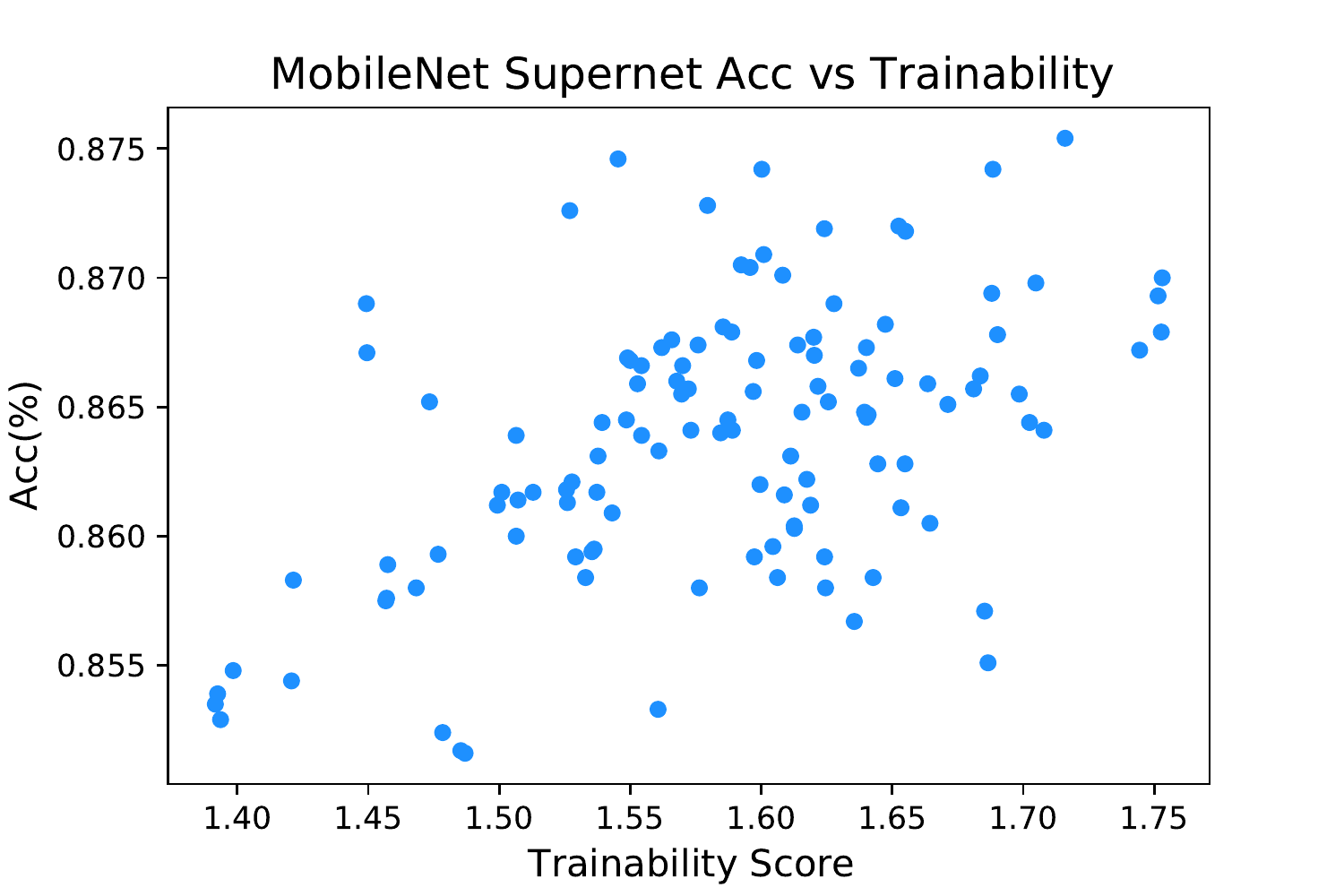}
     \end{subfigure}
     \caption{\textbf{Left}: The $\gamma$-indicator versus test accuracy on MobileNet supernet, \textbf{Right}: the trainability score versus test accuracy on MobileNet supernet}
     \label{fig:mobilenet_supernet_gamma}
\end{figure}
\begin{figure}[t]
     \centering
     \includegraphics[width=0.33\textwidth]{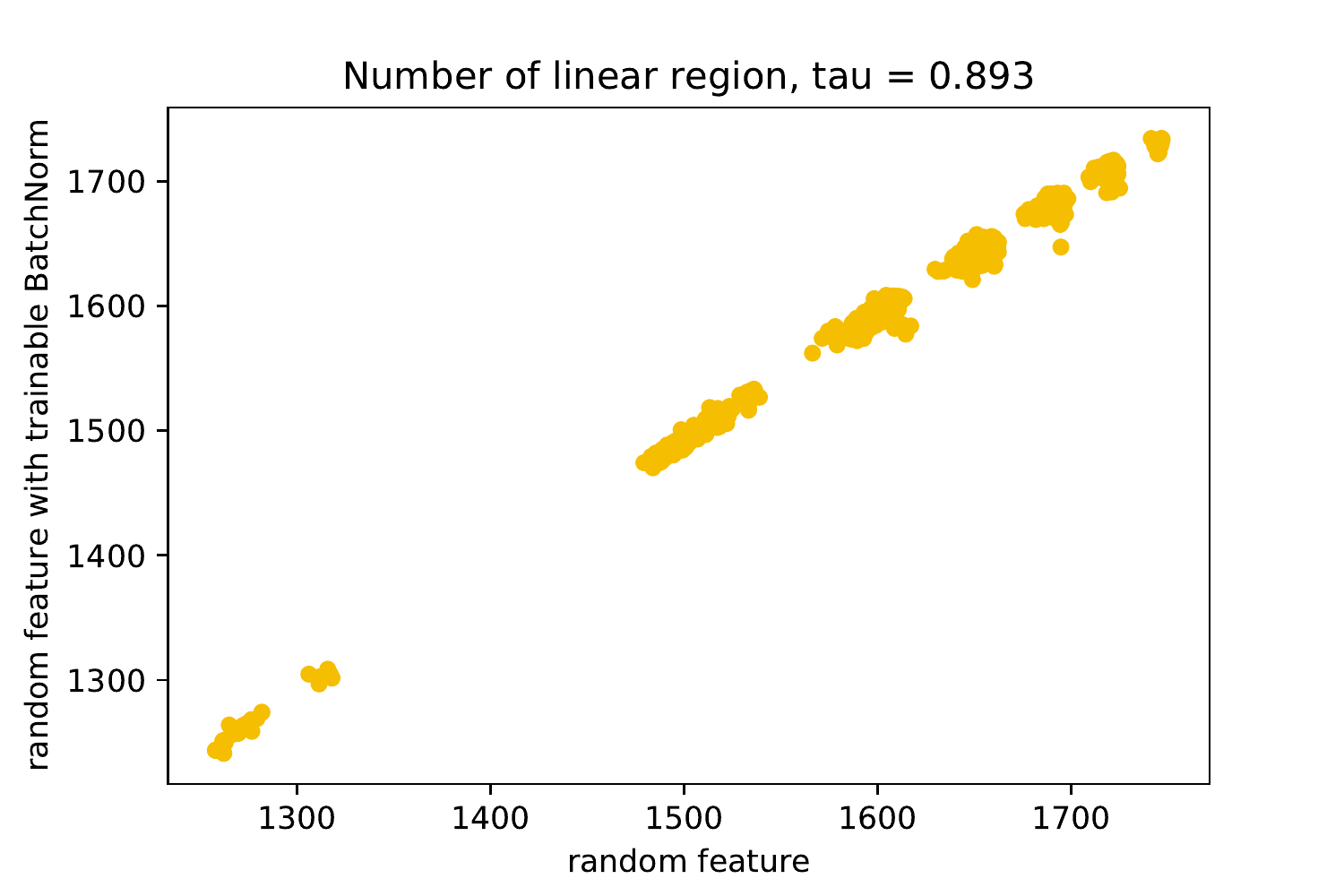}
     \caption{The number of linear regions does not changes as the BatchNorm are trained for more epochs.}
     \label{fig:mobilenet_supernet_gamma}
\end{figure}
\\
\\
\noindent
\textbf{Expressivity versus more epochs of training} We discuss the impact that training longer on FBN supernet has on architecture rank measured by expressivity. We randomly select 1,000 architectures from NAS-Bench-201. We compare the scores change after train for ten more epochs. We observe that the expressivity is consistent. Training more epochs on supernet does not result in a change in the architecture's rank. In conclusion, it is safe to measure the expressivity~\cite{mellor2021trainfreenas} of FBN supernet in the very beginning, thus saving lots of time on initialization.

\section{Conclusion}
In this work, we both theoretically and empirically study the train-BN-only strategy in NAS. First, we prove that the train-BN-only networks obtain the same training dynamics as the conventional optimization strategy does in networks by proving that optimizing infinite-width networks with updating BN-only is a neural tangent kernel. We then verify the unfairness in supernet training caused by the Conv-BN combination and propose a solution to solve the issue. Further, we leverage the theoretically-inspired BN properties from three perspectives to evaluate the networks. Overall, our method is general on diverse search space, fast on supernet training, and better at finding small and high-performing networks than the conventional one-shot NAS approach.

{\small
\bibliographystyle{ieee_fullname}
\bibliography{egbib}
}

\end{document}